\documentclass[11pt,a4paper]{article}
\usepackage[acceptedWithA]{tacl2018v2}
\usepackage{times,latexsym}
\usepackage{url}
\usepackage[T1]{fontenc}
\usepackage{CJKutf8}

\usepackage{xspace,mfirstuc,tabulary}

\usepackage{amsmath}
\usepackage{multirow}
\usepackage{comment}
\usepackage{xcolor}
\usepackage{ulem}
\makeatletter
\newcommand{\@emptybiblabel}[1]{}
\makeatother
\usepackage{color}
\usepackage{algorithm}
\usepackage{algpseudocode}
\usepackage{bold-extra}

\definecolor{myblue}{rgb}{0, 57, 230}
\definecolor{airforceblue}{rgb}{0.36, 0.54, 0.66}
\definecolor{navyblue}{rgb}{0, 0, 128}
\definecolor{ceruleanblue}{rgb}{0.16, 0.32, 0.75}
\definecolor{cornflowerblue}{rgb}{0.39, 0.58, 0.93}
\definecolor{denim}{rgb}{0.08, 0.38, 0.74}

\definecolor{azure(colorwheel)}{rgb}{0.0, 0.5, 1.0}
\definecolor{cornellred}{rgb}{0.7, 0.11, 0.11}

\definecolor{lemon}{rgb}{1.0, 0.97, 0.0}
\definecolor{amber(sae/ece)}{rgb}{1.0, 0.49, 0.0}
\definecolor{cadmiumorange}{rgb}{0.93, 0.53, 0.18}
\definecolor{darkorange}{rgb}{1.0, 0.55, 0.0}

\definecolor{debianred}{rgb}{0.84, 0.04, 0.33}
\definecolor{deepcarmine}{rgb}{0.66, 0.13, 0.24}
\definecolor{deepcarminepink}{rgb}{0.94, 0.19, 0.22}

\definecolor{brightgreen}{rgb}{0.4, 1.0, 0.0}
\definecolor{caribbeangreen}{rgb}{0.0, 0.8, 0.6}
\definecolor{chartreuse(web)}{rgb}{0.5, 1.0, 0.0}
\definecolor{darkpastelgreen}{rgb}{0.01, 0.75, 0.24}
\definecolor{electricgreen}{rgb}{0.0, 1.0, 0.0}
\definecolor{emerald}{rgb}{0.31, 0.78, 0.47}
\definecolor{cadmiumgreen}{rgb}{0.0, 0.42, 0.24}

\definecolor{darkmagenta}{rgb}{0.55, 0.0, 0.55}
\definecolor{darklavender}{rgb}{0.45, 0.31, 0.59}

\usepackage{helvet}  
\usepackage{courier}  
\usepackage{graphicx}  
\usepackage{amssymb}
\usepackage{amsfonts}
\usepackage{paralist}
\usepackage{bbold}
\usepackage{bm}
\usepackage[toc,page]{appendix} 
\usepackage{float}
\usepackage{makecell}
\usepackage{subfig}

\DeclareSymbolFontAlphabet{\amsmathbb}{AMSb}%

\newcommand{\R}[0]{\amsmathbb R}
\newcommand{\f}[0]{
\mathrm{Macro}\textrm{-}\mathrm{F}_1}

\DeclareMathOperator*{\softmax}{softmax}

\DeclareMathOperator*{\FFN}{FFN}
\DeclareMathOperator*{\SingleHead}{head}
\DeclareMathOperator*{\MultiHead}{multi-head}
\DeclareMathOperator*{\concat}{concat}

\DeclareMathOperator*{\where}{where}
\DeclareMathOperator*{\ef}{E-F}
\DeclareMathOperator*{\lf}{L-F}

\DeclareMathOperator*{\tp}{tp}
\DeclareMathOperator*{\fp}{fp}
\DeclareMathOperator*{\tn}{tn}
\DeclareMathOperator*{\fn}{fn}
\DeclareMathOperator*{\argmax}{arg\,max}

\usepackage{pict2e,picture}
\makeatletter
\DeclareRobustCommand{\Arrow}[1][]{%
\check@mathfonts
\if\relax\detokenize{#1}\relax
\settowidth{\dimen@}{$\m@th\rightarrow$}%
\else
\setlength{\dimen@}{#1}%
\fi
\sbox\z@{\usefont{U}{lasy}{m}{n}\symbol{41}}%
\begin{picture}(\dimen@,\ht\z@)
\roundcap
\put(\dimexpr\dimen@-.7\wd\z@,0){\usebox\z@}
\put(0,\fontdimen22\textfont2){\line(1,0){\dimen@}}
\end{picture}%
}
\makeatother
\newcommand{\veryshortrightarrow}{\hspace{.2mm}\scalebox{.8}{\Arrow[.1cm]}\hspace{.2mm}}

\newcommand{\mymodel}[0]{{\sc  DetNet}}

\newcommand{\sd}[0]{{\sc  DetNet$_{1\mathcal{H}}$}}
\newcommand{\wsd}[0]{{\sc  DetNet$_{2\mathcal{H}}$}}
\newcommand{\wsdprior}[0]{{\sc  DetNet$^\ast$}}

\newcommand{\detrank}[0]{\mbox{{\sc DetRank}}}
\newcommand{\textrank}[0]{\mbox{{\sc TextRank}}}

\newcommand{\mil}[0]{{\sc MilNet}}

\newcommand{\fsd}[0]{{\sc  \footnotesize DetNet$_{semi}$}}
\newcommand{\fwsd}[0]{{\sc  \footnotesize DetNet$_{post}$}}
\newcommand{\fwsdprior}[0]{{\sc  \footnotesize DetNet$^\ast$}}

\newcommand{\fmilsd}[0]{{\sc  MilNet}}
\newcommand{\fmilwsd}[0]{{\sc MilNet$_{w\veryshortrightarrow s\veryshortrightarrow d}$}}

\usepackage{cleveref}

\begin{document}
\title{Weakly Supervised Domain Detection}
 \author{Yumo Xu \and Mirella Lapata\\
 Institute for Language, Cognition and Computation\\
 School of Informatics, University of Edinburgh\\
 10 Crichton Street, Edinburgh EH8 9AB\\
 \texttt{yumo.xu@ed.ac.uk}, 
 \texttt{mlap@inf.ed.ac.uk}}
\date{}

\maketitle
\begin{abstract}
  In this paper we introduce \textit{domain detection} as a new
  natural language processing task. We argue that the ability to
  detect textual segments which are domain-heavy, i.e.,~sentences or
  phrases which are representative of and provide evidence for a given
  domain could enhance the robustness and portability of various text
  classification applications. We propose an \textit{encoder-detector}
  framework for domain detection and bootstrap classifiers with
  multiple instance learning (MIL). The model is hierarchically
  organized and suited to multilabel classification.  We demonstrate
  that despite learning with minimal supervision, our model can be
  applied to text spans of different granularities, languages, and
  genres.  We also showcase the potential of domain detection for text
  summarization.
\end{abstract}

\section{Introduction}
\label{sec:intro}

Text classification is a fundamental task in Natural Language
processing which has been found useful in a wide spectrum of
applications ranging from search engines enabling users to identify
content on websites, sentiment and social media analysis, customer
relationship management systems, and spam detection. Over the past
several years, text classification has been predominantly modeled as a
supervised learning problem
(e.g.,~\citealt{kim:2014:EMNLP2014,McCallum:Nigam:1998,iyyer-EtAl:2015:ACL-IJCNLP})
for which appropriately labeled data must be collected. Such data is
often domain-dependent (i.e.,~covering specific topics such as those
relating to ``Business'' or ``Medicine'') and a classifier trained
using data from one domain is likely to perform poorly on another. For
example, the phrase ``\textsl{the mouse died quickly}'' may indicate
negative sentiment in a customer review describing the hand-held
pointing device or positive sentiment when describing a laboratory
experiment performed on a rodent. The ability to handle a wide
variety of domains\footnote{The term ``domain'' has been permissively
  used in the literature to describe (a)~a collection of documents
  related to a particular topic such as user-reviews in Amazon for a
  product category (e.g.,~\emph{books, movies}), (b)~a type of
  information source (e.g., \emph{twitter, news articles}), and
  (c)~various fields of knowledge (e.g.,~\emph{Medicine, Law,
    Sport}). In this paper we adopt the latter definition of domains,
  however, nothing in our approach precludes applying it to different
  domain labels.} has become more pertinent with the rise of
data-hungry machine learning techniques like neural networks and their
application to a plethora of textual media ranging from news articles
to twitter, blog posts, medical journals, Reddit comments, and
parliamentary debates
\cite{kim:2014:EMNLP2014,Zichao:ea:2016,Conneau:ea:2017,Ye:et:al:2016}.

The question of how to best deal with multiple domains when training
data is available for one or few of them has met with much interest in
the literature. The field of \emph{domain adaptation}
\cite{jiang2007instance,blitzer2006domain,daume2007frustratingly,finkel2009hierarchical,lu2016general}
aims at improving the learning of a predictive function in a
\emph{target} domain where there is little or no labeled data, using
knowledge transferred from a \emph{source} domain where sufficient
labeled data is available. Another line of work
\cite{Li:Zong:2008,Wu:Huang:2015,Chen:Cardie:2018} assumes that
labeled data may exist for multiple domains, but in insufficient
amounts to train classifiers for one or more of them. The aim of
\emph{multi-domain text classification} is to leverage all the
available resources in order to improve system performance across
domains simultaneously.
%

In this paper we investigate the question of how domain-specific data
might be obtained in order to enable the development of text
classification tools as well as more domain aware applications such as
summarization, question answering, and information extraction.  We
refer to this task as \emph{domain detection} and assume a fairly
common setting where the domains of a corpus collection are known and
the aim is to identify textual segments which are domain-heavy,
i.e.,~documents, sentences, or phrases providing evidence for a given
domain.

Domain detection can be formulated as a \textit{multilabel}
classification problem, where a model is trained to recognize domain
evidence at the sentence-, phrase-, or word-level. By definition then,
domain detection would require training data with fine-grained domain
labels, thereby increasing the annotation burden; we must provide
labels for training domain detectors \emph{and} for modeling the task
we care about in the first place. In this paper we consider the
problem of fine-grained domain detection from the perspective of
\emph{Multiple Instance Learning} (MIL; \citealt{keeler1992self}) and
develop domain models with very little human involvement.  Instead of
learning from individually labeled segments, our model only requires
document-level supervision and optionally prior domain knowledge and
learns to introspectively judge the domain of constituent segments.
Importantly, we do not require document-level domain annotations
either since we obtain these via distant supervision by leveraging
information drawn from Wikipedia.

Our domain detection framework comprises two neural network modules;
an \emph{encoder} learns representations for words and sentences
together with prior domain information if the latter is available
(e.g.,~domain definitions), while a \emph{detector} generates
domain-specific scores for words, sentences, and documents. We obtain
a segment-level domain predictor which is trained end-to-end on
document-level labels using a hierarchical, attention-based neural
architecture \cite{vaswani2017attention}.  We conduct domain detection
experiments on English and Chinese and measure system performance
using both automatic and human-based evaluation.  Experimental results
show that our model outperforms several strong baselines and is robust
across languages and text genres, despite learning from weak
supervision. We also showcase our model's application potential for
text summarization.


Our contributions in this work are threefold; we propose domain
detection, as a new fine-grained multilabel learning problem which we
argue would benefit the development of domain aware NLP tools; we
introduce a weakly supervised \textit{encoder-detector} model within
the context of multiple instance learning; and demonstrate that it can
be applied across languages and text genres without modification.

\section{Related Work}
\label{sec:relwork}

Our work lies at the intersection of multiple research areas,
including domain adaptation, representation learning,  multiple
instance learning, and topic modeling. We review related work below.

\paragraph{Domain Adaptation} A variety of domain adaptation methods
\cite{jiang2007instance,arnold2007comparative,pan2010survey} have been
proposed to deal with the lack of annotated data in novel domains
faced by supervised models.  \newcite{daume2006domain} propose to
learn three separate models, one specific to the source domain, one
specific to the target domain, and a third one representing domain
general information.  A simple yet effective feature augmentation
technique is further introduced in \newcite{daume2007frustratingly}
which \newcite{finkel2009hierarchical} subsequently recast within a
hierarchical Bayesian framework.  More recently,
\newcite{lu2016general} present a general regularization framework for
domain adaptation while
\newcite{camacho2017babeldomains} integrate domain information within
lexical resources. A popular approach within text classification
learns features that are invariant across multiple domains whilst
explicitly modeling the individual characteristics of each domain
\cite{Chen:Cardie:2018,Wu:Huang:2015,Bousmalis:ea:2016}.


Similar to domain adaptation, our detection task also identifies the
most discriminant features for different domains. However, while
adaptation aims to render models more portable by transferring
knowledge, detection focuses on the domains themselves and identifies
the textual segments which provide the best evidence for their
semantics, allowing to create datasets with \emph{explicit} domain
labels to which domain adaptation techniques can be further applied.

\paragraph{Multiple Instance Learning} Multiple instance learning
(MIL) handles problems where labels are associated with groups or
\emph{bags} of instances (documents in our case), while instance
labels (segment-level domain labels) are unobserved.  The task is then
to make aggregate instance-level predictions, by inferring labels
either for bags
\cite{keeler1992self,dietterich1997solving,maron1998multiple} or
jointly for instances and bags
\cite{zhou2009multi,wei2014scalable,kotzias2015group}.  Our domain
detection model is an example of the latter variant.

Initial MIL models, adopted a relatively strong consistency assumption 
between bag labels and instance labels. For instance, in  binary
classification, a bag was considered positive only if all its instances were
positive
\cite{dietterich1997solving,maron1998multiple,zhang2002content,andrews2004multiple,carbonetto2008learning}.
The assumption was subsequently relaxed by 
investigating prediction combinations
\cite{weidmann2003two,zhou2009multi}.

Within NLP, multiple instance learning has been predominantly applied
to sentiment analysis. \newcite{kotzias2015group} use sentence vectors
obtained by a pre-trained hierarchical CNN \cite{Denil:ea:2014} as
features under a MIL objective which simply averages instance
contributions towards bag classification (i.e.,~positive/negative
document sentiment).  \newcite{pappas2014explaining} adopt a multiple
instance regression model to assign sentiment scores to specific
product aspects, using a weighted summation of predictions.  More
recently, \newcite{angelidis2018multiple} propose {\sc MilNet}, a
multiple instance learning network model for sentiment analysis. They
employ an attention mechanism to flexibly weigh predictions and
recognize sentiment-heavy text snippets (i.e.,~sentences or clauses).

We depart from previous MIL-based work, in devising an encoding module
with self-attention and non-recurrent structure, which is particularly
suitable for modeling long documents efficiently. Compared to {\sc
  MILNet} \cite{angelidis2018multiple}, our approach generalizes to
segments of \emph{arbitrary} granularity; it introduces an instance
scoring function which supports multilabel rather than binary
classification, and takes prior knowledge into account (e.g.,~domain
definitions) to better inform the model's predictions.

\paragraph{Topic Modeling} Topic models are built around the idea that
the semantics of a document collection is governed by latent
variables. The aim is therefore to uncover these latent
variables--topics--that shape the meaning of the document
collection. Latent Dirichlet Allocation (LDA;
\citealt{blei2003latent}) is one of the best-known topic models. In
LDA, documents are generated probabilistically using a mixture
over~$K$ topics which are in turn characterized by a distribution over
words. And words in a document are generated by repeatedly sampling a
topic according to the topic distribution and selecting a word given
the chosen topic.

Although most topic models are unsupervised, some variants can also
accommodate document-level supervision
\cite{mcauliffe2008supervised,lacoste2009disclda}.  However, these
models are not appropriate for analyzing multiply labeled corpora
since they limit documents to being associated with a single label.
Multi-Multinomial LDA (MM-LDA; \citealt{ramage2009clustering}) relaxes
this constraint by modeling each document as a bag of words with a bag
of labels, and topics for each observation are drawn from a shared
topic distribution.
Labeled LDA (L-LDA; \citealt{ramage2009labeled}) goes one step further
by directly associating labels with latent topics thereby learning
label-word correspondences.  L-LDA is a natural extension of both LDA
by incorporating supervision and Multinomial Naive Bayes
\cite{McCallum:Nigam:1998} by incorporating a mixture model
\cite{ramage2009labeled}.

Similar to L-LDA, \mymodel~is also designed to perform learning and
inference in multi-label settings.  Our model adopts a more general
solution to the credit attribution problem (i.e., the association of
textual units in a document with semantic tags or labels).
Despite learning from a weak and distant signal, our model can produce
domain scores for text spans of varying granularity (e.g.,~sentences
and phrases) not just words and achieves this with a
hierarchically-organized neural architecture. Aside from learning
through efficient backpropagation, the proposed framework can take
incorporate useful prior information (e.g.,~pertaining to the labels
and their meaning).

\section{Problem Formulation}
\label{sec:form}
\begin{figure}[t]
    \centering
   \includegraphics[width=7.5cm]{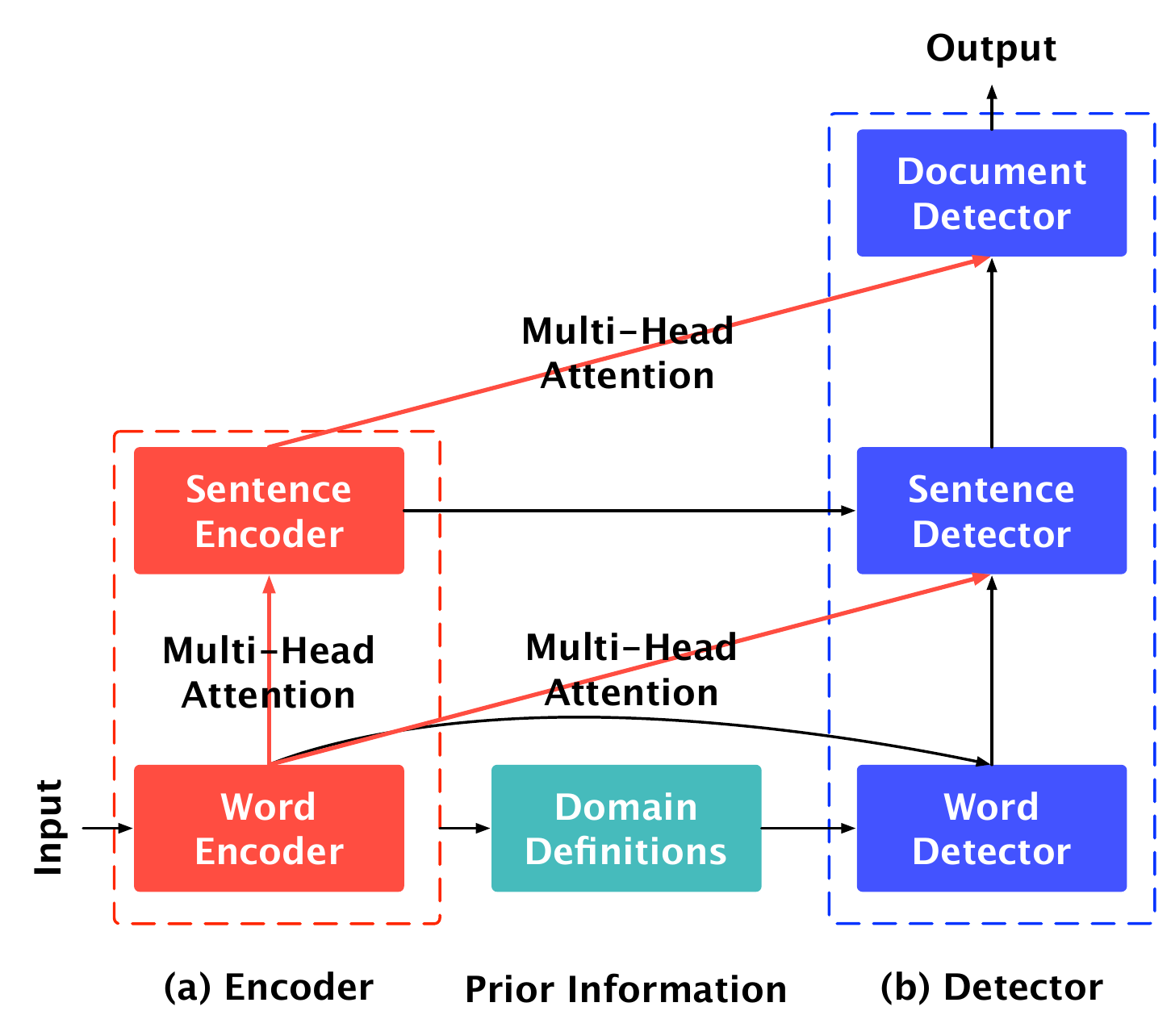}
   \caption{\label{fig:framework} Overview of \mymodel.  The encoder
     learns document representations in a hierarchical fashion and the
     decoder generates domain scores, whilst selectively attending to
     previously encoded information. Prior information can be
     optionally incorporated when available at the encoding stage
     through parameter sharing. }
\end{figure}

We formulate domain detection as a multilabel learning problem. 
Our model is trained on samples of document-label pairs. Each document
consists of~$s$ sentences $x=\{x_1, \dots, x_s\}$ and is associated
with  discrete labels $y=\{y^{(c)} \vert c \in \left[1, C\right]
\}$. In this work, domain labels are not annotated manually but
extrapolated from Wikipedia (see Section~\ref{sec:exp} for details).
In a non-MIL framework, a model typically learns to predict document
labels by directly conditioning on its sentence representations
$\bm{h}_1, \dots, \bm{h}_s$ or their aggregate.
In contrast, $y$~under MIL is a learned function~$f_{\theta}$ of
\emph{latent} instance-level labels, i.e.,~\mbox{$y = f_{\theta}(y_1,
  \dots, y_s)$}. A MIL classifier will therefore first produce domain
scores for all instances (aka sentences), and then learn to integrate
instance scores into a bag (aka document) prediction.

In this paper we further assume that the \textit{instance-bag}
relation applies to sentences and documents but also to words and
sentences.
In addition, we incorporate prior domain information to facilitate
learning in a weakly supervised setting: each domain is associated
with a \textit{definition}~$\mathcal U^{(c)}$, i.e.,~a few sentences
providing a high-level description of the domain at hand. For example,
the definition of the ``Lifestyle'' domain is ``the interests,
opinions, behaviors, and behavioral orientations of an individual,
group, or culture''.
Figure~\ref{fig:framework} provides an overview of our Domain
\textbf{Det}ection \textbf{Net}work, which we call \mymodel.  The
model comprises two modules; an \emph{encoder} learns
representations for words and sentences whilst incorporating prior
domain information; a \emph{detector} generates domain scores for
words, sentences, and documents by selectively attending to previously
encoded information. We describe the two modules in more detail below.

\section{The Encoder Module}
\label{sec:enc}
\begin{figure}[t]
    \centering
   \includegraphics[width=6cm]{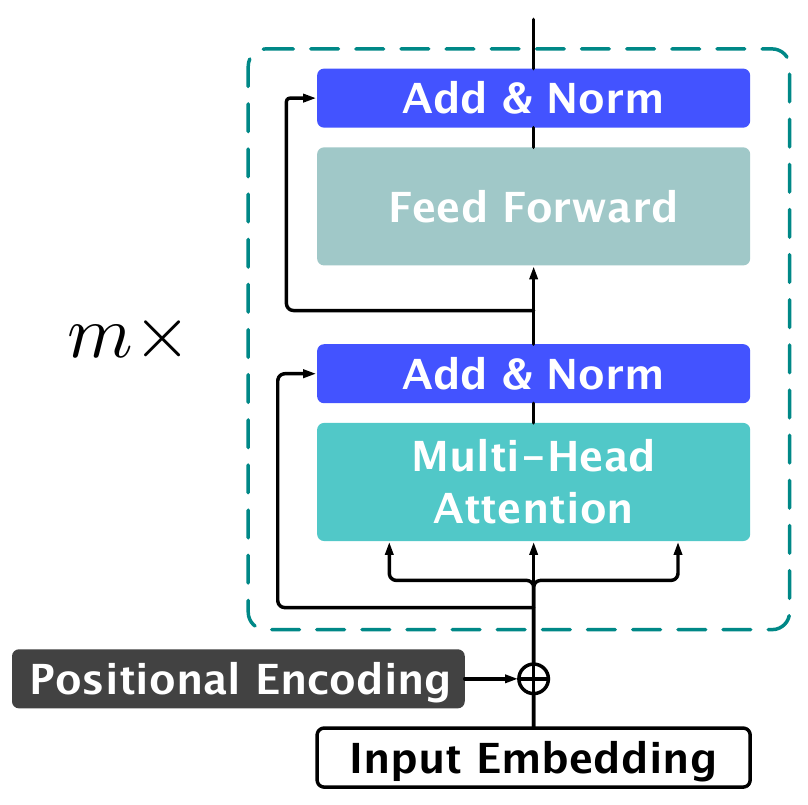}
    \caption{\label{fig:trans} Self-attentive encoder in Transformer
      \cite{vaswani2017attention} stacking $m$ identical layers.} 
\end{figure}

We learn representations for words and sentences using identical
encoders with separate learning parameters.  Given a document, the two
encoders implement the following steps:
\begin{align*}
    \bm{Z}, \bm{\alpha}= \textsc{WordEnc} (\bm{X}) \\
    \bm{G} = \left[ \bm{g}_1; \dots; \bm{g}_s\right] \where \;
    \bm{g} = \bm{Z} \bm{\alpha}\\
    \bm{H}, \bm{\beta}= \textsc{SentEnc} (\bm{G})
\end{align*}
For each sentence $\bm{X} = [ \bm{x}_1; \dots ; \bm{x}_n]$, the
word-level encoder yields contextualized word representations~$\bm{Z}$ and
their attention weights~$\bm{\alpha}$.  Sentence embeddings~$\bm{g}$
are obtained via weighted averaging and then provided as input to the
sentence-level encoder which outputs contextualized representations
$\bm{H}$ and their attention weights $\bm{\beta}$.

In this work we aim to model fairly long documents (e.g., Wikipedia
articles; see Section~\ref{sec:exp} for details). For this reason, our
encoder builds on the Transformer architecture
\cite{vaswani2017attention}, a recently proposed highly efficient
model which has achieved state-of-the-art performance in machine
translation \cite{vaswani2017attention} and question answering
\cite{yu2018qanet}. The Transformer aims at reducing the fundamental
constraint of sequential computation which underlies most
architectures based on recurrent neural networks. It eliminates
recurrence in favor of applying a self-attention mechanism which
directly models relationships between all words in a sentence,
regardless of their position.


\paragraph{Self-Attentive Encoder} As shown in Figure~\ref{fig:trans},
the Transformer is a non-recurrent framework comprising $m$~identical
layers. Information on the (relative or absolute) position of each
token in a sequence is represented by the use of positional encodings
which are added to input embeddings (see the bottom of
Figure~\ref{fig:trans}). We denote position-augmented inputs in a
sentence with~$\bm{X}$. Our model uses four layers in both word and
sentence encoders. The first three layers are identical to those in
the Transformer ($m=3$), comprising a multi-head self-attention
sublayer and a position-wise fully-connected feed-forward network. The
last layer is simply a multi-head self-attention layer yielding
attention weights for subsequent operations.

Single-head attention takes three parameters as input in the
Transformer \cite{vaswani2017attention}: a query matrix, a key matrix,
and a value matrix. These three matrices are identical and equal to
the inputs~$\bm{X}$ at the first layer of the word encoder. The output
of a single-head attention is calculated via:
\begin{equation}
    \SingleHead(\bm{X},\bm{X}, \bm{X}) = \softmax(\frac{\bm{X}\bm{X}^{\intercal}}{\sqrt{d_x}})\bm{X}.
\end{equation}


Multi-head attention allows to jointly attend to information from
different representation subspaces at different positions. This is
done by first applying different linear projections to inputs and then
concatenating them:
\begin{gather}
\hspace*{-1ex}{\SingleHead}^{(k)} = \SingleHead(\bm{XW}_1^{(k)},\bm{XW}_2^{(k)}, \bm{XW}_3^{(k)})\\
\hspace*{-2ex}\MultiHead\hspace*{-0.4ex}=\hspace*{-0.4ex} \concat({\SingleHead}^{(1)},.., {\SingleHead}^{(r)})\bm{W}_4
\end{gather}
where we adopt four heads ($r=4$) for both word and sentence encoders.
The second sublayer in the Transformer (see Figure~\ref{fig:trans}) is
a fully-connected feed-forward network applied to each position
separately and identically:\footnote{We omit here the bias term for
  the sake of simplicity.}
\begin{equation}
    \FFN(\bm{x}) = \max(0, \bm{xW}_5)\bm{W}_6 .
\end{equation}

After sequentially encoding input embeddings through the first three
layers, we obtain contextualized word representations $\bm{Z} \in
\R^{d_z \times n }$.  Based on~$\bm{Z}$, the last multi-head attention
layer in the word encoder yields a set of attention matrices $\mathcal
A = \{\bm{A}^{(k)}\}_{k=1}^{r}$ for each sentence where $\bm{A}^{(k)}
\in \R^{n \times n }$.
Therefore, when measuring the contributions from words to sentences,
e.g., in terms of domain scores and representations, we can
selectively focus on salient words within the set~$\mathcal A =
\{\bm{A}^{(k)}\}_{k=1}^{r}$:
\begin{equation}
    \label{eq:alpha}
    \bm{\alpha} = \softmax(\frac{1}{\sqrt{n r}} \sum^{r}_k \sum^{n}_{\ell} \bm{A}^{(k)}_{\star, \ell})
\end{equation}
where the $\softmax$ function outputs the salience distribution over words:
\begin{equation}
    \softmax (\bm{a}_\ell) = \frac{e^{\bm{a}_\ell}}{\sum_{\bm{a}_{\ell'} \in \bm{a}} e^{\bm{a}_{\ell'}}}
\end{equation}
and obtain sentence embeddings $\bm{g} = \bm{Z} \bm{\alpha}$. 

In the same vein, we adopt another self-attentive encoder to obtain
contextualized sentence representations
$\bm{H} \in \R^{d_h \times s}$. The final layer outputs multi-head
attention score matrices $\mathcal B = \{\bm{B}^{(k)}\}_{k=1}^{r}$
(with $\bm{B}^{(k)}\in \R^{s \times s}$), and we calculate sentence
salience as:
  \begin{equation}
  \label{eq:beta}
    \bm{\beta} = \softmax(\frac{1}{\sqrt{sr}} \sum^r_k \sum^s_j \bm{B}^{(k)}_{\star, j}).
    \end{equation}



\noindent{\bf Prior Information} 
In addition to documents (and their domain labels), we might have some
prior knowledge about the domain, e.g., its general semantic content
and the various topics related to it. For example, we might expect
articles from the ``Lifestyle'' domain to not talk about missiles
or warfare, as these are recurrent themes in the ``Military''
domain. As mentioned earlier, throughout this paper we assume we have
domain definitions~$\mathcal U$ expressed in a few sentences as prior
knowledge.
Domain definitions share parameters with \textsc{WordEnc} and
\textsc{SentEnc} and are encoded in a \textit{definition}
matrix~$\bm{U} \in \R^{d_h \times C}$. 

Intuitively, identifying the domain of a word might be harder than
that of a sentence; on account of being longer and more expressive,
sentences provide more domain-related cues than words whose meaning
often relies on supporting context.  We thus inject domain
definitions~$\bm{U}$ into our word detector only.

\section{The Detector Module}

\label{sec:det}
\begin{figure}[t]
    \centering
    \includegraphics[width=8cm]{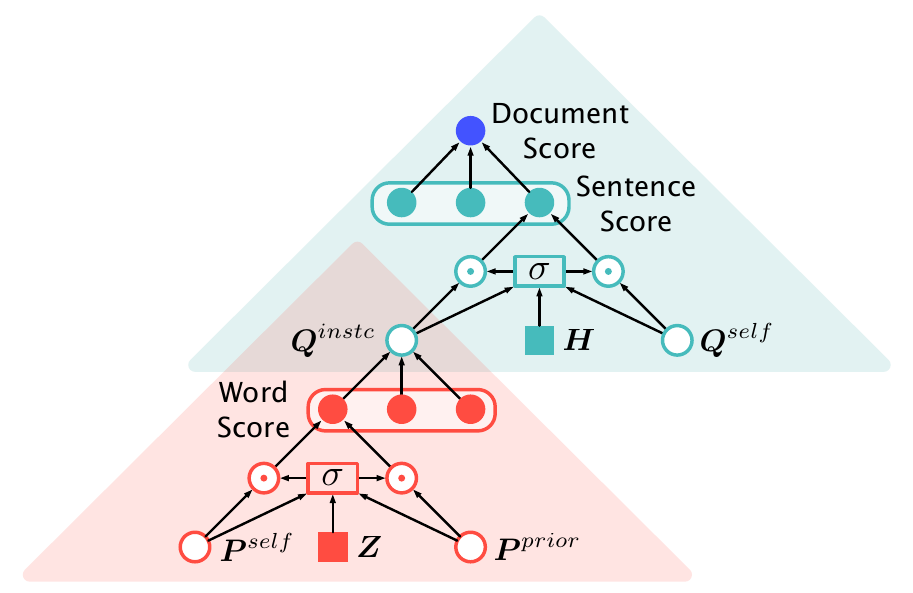}
    \caption{\label{fig:gates} Domain predictions for words and
      sentences;  the instance-bag relation applies
      to words-sentences (red shadow) and  sentences-documents (green
      shadow).     Squares denote representations of words or
      sentences, and circles 
are  domain scores.  
  }
\end{figure}

\textsc{DetNet} adopts three detectors corresponding to words,
sentences, and documents:
\begin{align*}
    \bm{P}= \textsc{WordDet} (\bm{Z}, \bm{U}) \\
    \bm{Q}^{instc} = \left[\bm{q}^{instc}_1; \dots; \bm{q}^{instc}_s\right] \where\;
    \bm{q}^{instc} = \bm{P} \bm{\alpha}\\
    \bm{Q}= \textsc{SentDet} (\bm{Q}^{instc}, \bm{H}) \\
    \tilde{\bm{y}} = \textsc{DocDet} (\bm{Q}, \bm{\beta})
\end{align*}
\textsc{WordDet} first produces word domain scores using both
lexical semantic information~$\bm{Z}$ and prior (domain)
knowledge~$\bm{U}$; \textsc{SentDet} yields domain scores for
sentences while integrating downstream instance
signals~$\bm{Q}^{instc}$ and sentence semantics~$\bm{H}$; finally,
\textsc{DocDet} makes the final document-level predictions based on
sentence scores.


\paragraph{Word Detector} Our first detector yields word domain
scores.  For a sentence, we obtain a \textit{self-scoring}
matrix~$\bm{P}^{self}$ using its own contextual word semantic
information:
\begin{equation}
\label{eq:p_self}
    \bm{P}^{self} = \tanh (\bm{W}_z \bm{Z}).
\end{equation}
%
In contrast to the representations used in
\newcite{angelidis2018multiple}, we generate instance scores from
contextualized representations, i.e.,~$\bm{Z}$.  Since the $\softmax$
function normally favors single-mode outputs, we adopt $\tanh(\cdot)
\in (-1, 1)$ as our domain scoring function to tailor MIL to our
multilabel scenario.

As mentioned earlier, we employ domain definitions as prior
information at the word level and compute the \textit{prior} score via:
\begin{equation}
\label{eq:p_prior}
    \bm{P}^{prior} = \tanh(\max(\bm{0}, \bm{U} ^\intercal \bm{W}_{u})  \bm{Z})
\end{equation}
where $\bm{W}_u \in \R^{d_u \times d_z}$ projects prior information
$\bm{U}$ onto the input semantic space. The prior score matrix
$\bm{P}^{prior}$ captures the interactions between domain definitions
and sentential contents.

In this work, we flexibly integrate scoring components with gates, as
shown in Figure~\ref{fig:gates}. The key idea is to learn a
\emph{prior gate}~$\bm{\Gamma}$ balancing
Equations~(\ref{eq:p_self}) and~(\ref{eq:p_prior}) via:
\begin{gather}
    \bm{\Gamma} = \gamma \sigma(\bm{W}_{g, p} [\bm{Z}, \bm{P}^{self}, \bm{P}^{prior}]) \label{eq:p_prior_gate}\\
    \bm{P} = \bm{\Gamma} \odot \bm{P}^{prior} + (\bm{J}-\bm{\Gamma})  \odot \bm{P}^{self} \label{eq:p}
\end{gather}
where $\bm{J}$ is an all-ones matrix and $\bm{P} \in \R^{C \times n}$
is the final domain score matrix at the word-level; $\odot$~denotes
element-wise multiplication and~$[\cdot, \cdot]$ matrix concatenation.
$\sigma(\cdot) \in (0, 1)$ is the sigmoid function and
$\bm{\Gamma} \in (0, \gamma)$ the prior gate with scaling
factor~$\gamma$, a hyperparameter controlling the overall effect of
prior information and instances.\footnote{Initially, we expected to
  balance these effects by purely relying on the learned function
  without a scaling factor. This, however, led to poor performance.}




\paragraph{Sentence Detector}

The second detector identifies sentences with domain-heavy semantics
based on signals from the sentence encoder, prior information and word
instances. Again we obtain a self-scoring matrix~$\bm{Q}^{self}$ via:
\begin{equation}
\label{eq:q_self}
     \bm{Q}^{self} = \tanh(\bm{W}_h \bm{H}).
\end{equation}

After computing sentence scores from sentence-level signals, 
we estimate domain scores from individual words.  We do this by reusing
$\bm{\alpha}$ in Equation~(\ref{eq:alpha}), $\bm{q}^{instc} = \bm{P}\bm{\alpha}$. After gathering $\bm{q}^{instc}$ for each sentence, we
obtain $\bm{Q}^{instc} \in \R^{C \times s}$ as the full instance score
matrix.

Analogously to the word-level detector (see Equation~(\ref{eq:p_prior_gate})), we employ  a sentence-level
\textit{upward gate} $\bm{\Lambda}$ to dynamically propagate domain scores from downstream word instances
to sentence bags:
\begin{gather}
      \bm{\Lambda} = \lambda \sigma(\bm{W}_{\ell} [\bm{H}, \bm{Q}^{instc}, \bm{Q}^{self}]) \\
     \bm{Q} =\bm{\Lambda} \odot \bm{Q}^{instc} +  (\bm{J} - \bm{\Lambda})  \odot \bm{Q}^{self}
\end{gather}
where $\bm{Q}$ is the final sentence score matrix.


\paragraph{Document Detector} 
Document-level domain scores are based on the sentence salience
distribution $\bm{\beta}$ (see Equation (\ref{eq:beta})) and are
computed as the weighted average of sentence scores:
\begin{equation}
\tilde{\bm{y}} = \bm{Q} \bm{\beta}.
\end{equation}

We use only document-level supervision for multilabel learning in $C$
domains. Formally, our training objective is:
\begin{equation}
    \mathcal L = 
    \min -\frac{1}{N} \sum_i^N \sum_c^C 
\log ( 1 + e^{-\tilde{\bm{y}}^{(i)}_c  \bm{y}^{(i)}_{c}})
\end{equation}
where $N$ is the training set size. At test time, we partition domains
into a relevant set and an irrelevant set for unseen samples.  Since
the domain scoring function is $\tanh(\cdot) \in (-1, 1)$, we use a
threshold of $0$ against which $\bm{\tilde y}$ is
calibrated.\footnote{If $\forall c \in [1, C]: \bm{\tilde y}_c < 0$
  holds, we set
  $\bm{\tilde y}_{c^\ast} = 1$ and select $c^\ast$ as $c^\ast = \argmax_c{\bm{\tilde
      y}_{c}}$ to produce a positive prediction.}

\section{Experimental Setup}
\label{sec:exp}
\begin{table}[t]
\centering
\begin{tabular}{|@{~}l| r r@{~}|}  
\hline
& \textbf{Wiki-en} & \textbf{Wiki-zh}\\
\hline \hline
All Documents & 31{,}562 & 26{,}280\\
Training Documents & 25{,}562 & 22{,}280\\
Development Documents & 3{,}000 & 2{,}000\\
Test Documents & 3{,}000 & 2{,}000\\
Multilabel Ratio & 10.18\%  & 29.73\%  \\
Average \#Words & 1{,}152.08 & 615.85\\
Vocabulary Size & 175{,}555 & 169{,}179\\
\hline \hline
Synthetic Documents & 200 & 200\\
Synthetic Sentences & 18{,}922 & 18{,}312 \\
\hline
\end{tabular}
\caption{\label{tab:data_stats} Statistics of Wikipedia datasets; en
  and zh are shorthands for English and Chinese, respectively. Synthetic
  documents and sentences are used in our automatic evaluation
  experiments discussed in Section~\ref{sec:sent_eval}.} 
\end{table}

\begin{algorithm}[t]
\small
\caption{Document Generation}
\label{alg:doc_syn}
\textbf{Input:} 
$\mathcal S = \{\mathcal S_d\}_1^D$: Label combinations\\
\makebox[1.05cm]{} $\mathcal O = \{\mathcal O_d\}_1^D$: Sentence subcorpora\\
\makebox[1.05cm]{} $\ell_{max}$: Maximum document length

\textbf{Output:} A synthetic document
\begin{algorithmic}[0]
\Function{Generate}{$\mathcal S, \mathcal O, \ell_{max}$}
    \State Generate a document domain set $ \mathcal S^{doc} \in \mathcal S$
    \State $\mathcal S^{sent} \gets \mathcal S^{doc} \cup \{\texttt{GEN}\}$
    
    \If{$\lvert \mathcal S^{sent} \rvert < C$ } \Comment{Number of domain labels}
        \State Generate a noisy domain $\epsilon \in \mathcal Y \setminus \mathcal S^{sent}$
        \State $\mathcal S^{sent} \gets \mathcal S^{sent} \cup \{\epsilon\}$
    \EndIf

   \State $\mathcal S^{cdt} \gets \emptyset$; \Comment{A set of candidate domain sets}
   
   \For{$\mathcal S_d \in \mathcal S$}
           \If{$\mathcal S_d \in \mathcal S^{sent}$}
           \State $\mathcal S^{cdt} \gets  \mathcal S^{cdt} \cup \{\mathcal S_d\}$
           \EndIf
       \EndFor
           
    \State $n_{label} \gets \lvert \mathcal S^{cdt} \rvert$\Comment{Number of unused labels}
    \State $n_{sent} \gets \ell_{max}$ \Comment{Number of sentence blocks}
    \State $\mathcal L \gets \emptyset$ \Comment{For generated sentences}

    \For{$\mathcal S_d^{cdt} \in \mathcal S^{cdt}$}
        \State $\theta = \min(\lvert \mathcal O_d \rvert, n_{sents} +1 -n_{labels}, \frac{2n_{sents}}{n_{labels}})$
        \State Generate $\ell_d \sim Uniform (1, \theta)$
        \State Generate $\ell_d$ sentences $\mathcal L_d \subseteq \mathcal O_d$ 
        \State $\mathcal L \gets  \mathcal L \cup \mathcal L_c$
        \State $n_{sent} \gets  \ell_{max} - |\mathcal L|$
        \State $n_{label} \gets  n_{label} -1$
    \EndFor
\State $\mathcal L \gets$ \Call{Shuffle}{$\mathcal L$}
\State \Return $\mathcal L$
\EndFunction
\end{algorithmic}
\end{algorithm}

\paragraph{Datasets} \textsc{DetNet} was trained on two datasets
created from
Wikipedia\footnote{http://static.wikipedia.org/downloads/2008-06/en}
for English and Chinese.\footnote{Available at https://github.com/yumoxu/detnet} Wikipedia articles are organized according
to a hierarchy of categories representing the defining characteristics
of a field of knowledge.  We recursively collect Wikipedia pages by
first determining the \textit{root categories} based on their match
with the domain name. We then
obtain their subcategories, the subcategories of these
subcategories, and so on. We treat all pages associated with a category
as representative of the domain of its root category.  

In our experiments we used seven target domains: {``Business and
  Commerce''} (\texttt{BUS}), {``Government and Politics''}
(\texttt{GOV}), {``Physical and Mental Health''} (\texttt{HEA}),
{``Law and Order''} (\texttt{LAW}), {``Lifestyle''} (\texttt{LIF}),
{``Military''} (\texttt{MIL}), and {``General Purpose''}
(\texttt{GEN}).  Exceptionally, \texttt{GEN} does not have a natural
root category. We leverage Wikipedia's 12~Main
Categories\footnote{https://en.wikipedia.org/wiki/Portal:Contents/Categories}
to ensure that \texttt{GEN} is genuinely different from the other six
domains. We used 5,000 pages for each domain.
Table~\ref{tab:data_stats} shows various statistics on our dataset. 

\label{sec:model_comp}
\paragraph{System Comparisons}
We constructed three variants of \mymodel~to explore the contribution
of different model components.  \sd~has a single-level hierarchical
structure, treating only sentences as instances and documents as bags;
while \wsd~ has a two-level hierarchical structure (the instance-bag
relation applies to words-sentences and sentences-documents); finally,
\wsdprior~is our full model which is fully hierarchical and equipped
with prior information (i.e.,~domain definitions).  We also compared
\mymodel~to a variety of related systems which include:

{\sc \textbf{Major}}: the  Majority domain label applies  to all instances. 

\textbf{L-LDA}: Labeled LDA \cite{ramage2009labeled} is a topic model
that constrains LDA by defining a one-to-one correspondence between
LDA's latent topics and observed labels. This allows L-LDA to directly
learn word-label correspondences. We obtain domain scores for words
through the topic-word-count matrix $\bm{M} \in \R^{C \times V}$ which
is computed during training:
\begin{equation}
\label{eq:llda_count_mat}
\tilde{\bm{M}} = \frac{\bm{M}^\intercal + \beta}{\sum^C_c \bm{M}^\intercal_{*,c} + C * \beta}
\end{equation}
where $C$ and $V$ are the number of domain labels and the size of
vocabulary, respectively. Scalar $\beta$ is a prior value set to $1/C$
and matrix $\tilde{\bm{M}} \in \R^{V \times C}$ consists of word
scores over domains. Following the snippet extraction approach
proposed in \newcite{ramage2009labeled}, L-LDA can also be used to
score sentences as the expected probability that the domain label had
generated each word. For more details on L-LDA, we refer the interested
reader to \newcite{ramage2009labeled}.

{\sc \textbf{HierNet}}: A hierarchical neural network model described
in \newcite{angelidis2018multiple} which produces document-level
predictions by attentively integrating sentence representations. For
this model we used word and sentence encoders identical to
\textsc{DetNet}. \textsc{HierNet} does not generate instance-level
predictions, however, we assume that document-level predictions apply
to all sentences. 

\textsc{\textbf{MilNet}}: 
A variant of the
MIL-based model introduced in \citet{angelidis2018multiple} which considers
sentences as instances and documents as bags (while \textsc{DetNet}
generalizes the instance-bag relationship to words and sentences).  To
make \mil~comparable to our system, we use an encoder identical to \mymodel, i.e., two Transformer encoders for words and sentences, respectively. Thus, \mil~differs from \sd~in two respects: (a)~word representations are simply averaged without
word-level attention to build sentence embeddings and (b) context-free
sentence embeddings generate sentence domain scores before being fed
to the sentence encoder.



\paragraph{Implementation Details} We used 16~shuffled samples in a
batch where the maximum document length was set to 100 sentences with the excess clipped. Word embeddings were initialized randomly with~256
dimensions. All weight matrices in the model were initialized with the
fan-in trick \cite{glorot2010understanding} and biases were initialized
with zero.  Apart from using layer normalization \cite{ba2016layer} in
the encoders, we applied batch normalization \cite{ioffe2015batch} and a
dropout rate of $0.1$ in the detectors to accelerate model training.
We trained the model with the Adam optimizer \cite{KingmaB14}. 
We set all three gate
scaling factors in our model to~$0.1$.  Hyper-parameters were optimized
on the development set.  To make our experiments easy to replicate, we
release our PyTorch \cite{paszke2017automatic} source
code.\footnote{Available at {https://github.com/yumoxu/detnet}}

\section{Automatic Evaluation}
\label{sec:sent_eval}

In this section we present the results of our automatic evaluation for
sentence and document predictions. Problematically, for sentence
predictions we do not have gold-standard domain labels (we have only
extrapolated these from Wikipedia for documents). We therefore
developed an automatic approach for creating silver standard domain
labels which we describe below. 

\paragraph{Test Data Generation} In order to obtain sentences with
domain labels, we exploit \emph{lead} sentences in Wikipedia
articles. Lead sentences typically define the article's subject matter
and emphasize its topics of
interest.\footnote{https://en.wikipedia.org/wiki/Lead\_paragraph} As
most lead sentences contain domain-specific content we can fairly
confidently assume that document-level domain labels will apply.  To
validate this assumption, we randomly sampled~20 documents
containing~220 lead sentences and asked two annotators to label these
with domain labels. Annotators overwhelmingly agreed in their
assignments with the document labels, the (average) agreement was
$K=0.89$ using Cohen’s Kappa coefficient.


We used the lead sentences to create pseudo documents simulating real
documents whose sentences cover multiple domains. To ensure sentence
labels are combined reasonably (e.g.,~\texttt{MIL} is not likely to
coexist with \texttt{LIF}), prior to generating synthetic documents,
we traverse the training set and acquire all domain combinations
$\mathcal S$,
e.g.,~$\mathcal S = \{\{\texttt{GOV}\},\{\texttt{GOV},
\texttt{MIL}\}\}$. We then gather lead sentences representing the same domain
combinations. We generate synthetic documents with a maximum length of~100
sentences (we also clip real documents to the same length). 

Algorithm~1 shows the pseudocode for document generation.  We first
sample document labels, then derive candidate label sets for sentences
by introducing \texttt{GEN} and a noisy label $\epsilon$. 
After sampling sentences for each
domain, we shuffle them to achieve domain-varied sentence contexts.
We created two synthetic datasets for English and Chinese.  Detailed
statistics are shown in Table~\ref{tab:data_stats}.

\paragraph{Evaluation Metric} We evaluate system performance
automatically using label-based $\f$ \cite{zhang2014review}, a
widely-used metric for multilabel classification.  It measures model
performance for each label specifically and then macro-averages
the results. For each class, given a confusion matrix
$\bigl(\begin{smallmatrix} {\tp} & {\fn} \\ {\fp} &
  {\tn} \end{smallmatrix}\bigr)$
containing the number of samples classified as true positive, false
positive, true negative, and false negative, $\f$ is calculated as
$\frac{1}{C}\sum_{c=1}^C\frac{2\tp_{c}}{2\tp_{c}+\fp_{c}+\fn_{c}}$
where $C$ is the number of domain labels.

\begin{table}[t]
\centering
\tabcolsep=0.15cm
\begin{tabular}{|l |l l| l l|}  
\hline
\multirow{2}*{\textbf{Systems}} &\multicolumn{2}{c}{\textbf{Sentences}}
& \multicolumn{2}{c|}{\textbf{Documents}}\\
~ & \multicolumn{1}{c}{en} & \multicolumn{1}{c|}{zh} & \multicolumn{1}{c}{en} & \multicolumn{1}{c|}{zh}\\
\hline  \hline
{\sc  Major}   & \hspace*{1.14ex}$2.81^{\dagger}$ & \hspace*{1.14ex}$5.99^{\dagger}$ &
\hspace*{1.15ex}$3.81^{\dagger}$ & \hspace*{1.15ex}$4.41^{\dagger}$\\
L-LDA  & $38.52^{\dagger}$ & $37.09^{\dagger}$ & $63.10^{\dagger}$  & $58.74^{\dagger}$\\
{\sc HierNet}  & $30.01^{\dagger}$ & $37.26^{\dagger}$ & $75.00$   & $68.56^{\dagger}$\\
\fmilsd & $37.12^{\dagger}$& $44.37^{\dagger}$ & $50.90^{\dagger}$& $69.45^{\dagger}$\\
\hline \hline
\sd  &  $47.93^{\dagger}$ & $51.31^{\dagger}$ & $74.91$
&  $72.85$\\
\wsd &  $47.89^{\dagger}$ & $52.50^{\dagger}$ & $75.47$
& $71.96^{\dagger}$ \\
\wsdprior &  $\bm{54.37}$ & $\bm{55.88}$ & $\bm{76.48}$ &  $\bm{74.24}$\\
\hline
\end{tabular}
\caption{\label{tab:doc_sent} 
  Performance using $\f\%$ on automatically created Wikipedia
  test set;   models with the symbol~$\dagger$ are significantly (\mbox{$p <
    0.05$}) different from the best  system in each task using
  the approximate randomization test \cite{noreen1989computer}.} 
\end{table}

\paragraph{Results} Our results are summarized in
Table~\ref{tab:doc_sent}.  
We first report domain detection results for documents, since reliable
performance on this task is a prerequisite for more fine-grained
domain detection.  As shown in Table~\ref{tab:doc_sent},
\textsc{DetNet} does well on document-level domain detection,
managing to outperform systems over which it has no clear advantage
(such as \textsc{HierNet} or \textsc{MilNet}). 

As far as sentence-level prediction is concerned, all {\sc DetNet}
variants significantly outperform all comparison systems. Overall,
\wsdprior~is the best system achieving~54.37\% and~55.88\% $\f$ on
English (en) and Chinese (zh), respectively. It outperforms \mil~by~17.25\% on English and~11.51\% on Chinese.  The
performance of the fully hierarchical model \wsd~is better than \sd,
showing positive effects of directly incorporating word-level domain
signals.
We also observe that prior information is generally helpful on both
languages and both tasks.





\section{Human Evaluation} 
\label{sec:human_eval}
\begin{table*}[t]
\bgroup
\def\arraystretch{1.0}
\centerline{
\begin{tabular}{|l| c c c |c c c|}  
\hline
\multirow{2}*{\textbf{Systems}} &\multicolumn{3}{c}{\textbf{Sentences}}
& \multicolumn{3}{c|}{\textbf{Words}}\\
~ & {Wiki-en} & {Wiki-zh} & {NYT} & {Wiki-en} & {Wiki-zh} & {NYT} \\
\hline  \hline
{\sc  Major}  &  \hspace*{1.13ex}$1.34^{\dagger}$ & \hspace*{1.13ex}$6.14^{\dagger}$ & \hspace*{1.13ex}$0.51^{\dagger}$ & \hspace*{2.5ex}$1.39^{\dagger}$ & $14.95^{\dagger}$  & \hspace*{1ex}$0.39^{\dagger}$\\
L-LDA  & $27.81^{\dagger}$ & $28.94^{\dagger}$ & $28.08^{\dagger}$ &  \hspace*{1.13ex}$24.58^{\dagger}$ & \hspace*{-1ex}$42.67$ & \hspace*{-.8ex}$26.24$  \\
{\sc HierNet}  & $42.23^{\dagger}$  & $29.93^{\dagger}$ & $44.74^{\dagger}$ & \hspace*{1.2ex}$15.57^{\dagger}$ 	& $24.25^{\dagger}$ 	& $18.27^{\dagger}$ \\
{\sc MilNet} & $39.30^{\dagger}$& $45.14^{\dagger}$ &  $29.31^{\dagger}$ & \hspace*{1.2ex}$22.11^{\dagger}$ 	& $33.10^{\dagger}$ 	& $23.33^{\dagger}$  \\
\hline \hline
\sd  & $48.12^{\dagger}$ &  $51.76^{\dagger}$ & $57.06^{\dagger}$ & \hspace*{1.2ex}$16.21^{\dagger}$ 	& $26.90^{\dagger}$ &	$21.61^{\dagger}$ \\
\wsd  & $54.70^{\dagger}$ &  $\bm{57.60}$ & $55.78^{\dagger}$ & \hspace*{.2ex}$\bm{27.06}$ & \hspace*{-1ex}$\bm{43.82}$ & 	\hspace*{-1ex}$26.52$ \\
\wsdprior & $\bm{58.01}$ &   $51.28^{\dagger}$ & $\bm{60.62}$ & $26.08$ 	& \hspace*{-1ex}$43.18$ 	& \hspace*{-1ex}$\bm{27.03}$ \\
\hline
\end{tabular}
}
\egroup
\caption{\label{tab:human_eval} 
  System performance using $\f\%$ (test set created via AMT); models
  with the symbol~$\dagger$ are significantly (\mbox{$p < 0.05$}) different from the best system in
  each task using the approximate  randomization test \cite{noreen1989computer}.}
\end{table*}

\begin{table}[t]
\centering
\begin{tabular}{|@{~}c|ccc|} \hline
\textbf{Domains} & \textbf{Wiki-en} & \textbf{Wiki-zh} & \textbf{NYT} \\\hline\hline
\texttt{BUS}& $78.65$&$68.66$&$77.33$\\
\texttt{HEA}& $42.11$&$81.36$&$64.52$\\
\texttt{GEN}& $43.33$&$37.29$&$43.90$\\
\texttt{GOV}& $80.00$&$37.74$&$62.07$\\
\texttt{LAW}& $69.77$&$41.03$&$46.51$\\
\texttt{LIF}& $17.24$&$27.91$&$50.00$\\ 
\texttt{MIL}& $75.00$&$65.00$&$80.00$\\\hline \hline
Avg &  $58.01$ & $51.28$     & $60.62$ \\\hline
\end{tabular}
\caption{\label{tab:performance_over_doms} Sentence-level \wsdprior~performance
  ($\f\%$) across domains on three datasets.}
\end{table}

\begin{table*}[t]
\small
\centering
\bgroup
\def\arraystretch{1.0}
\begin{tabular}{|@{~}c@{~}|@{~}p{7.3cm}@{~}|@{~}p{7cm}@{~}|}\hline
\textbf{Domains} &  \multicolumn{1}{@{~}c@{~}|@{~}}{\wsdprior} & \multicolumn{1}{c|}{L-LDA} \\\hline\hline
\texttt{BUS}&  monopolization, enactment, panama, funding, 
arbitron, maturity, groceries, os, 
elevator, salary, organizations, pietism,
 contract, mercantilism, sectors & 
also, business, company, used, one,
management, may, business, united, 
2007, time, first, new, market, new \\\hline
\texttt{HEA}& 
psychology, divorce, residence, pilates, 
dorlands, culinary, technique, emotion, 
affiliation, seafood, famine, malaria, 
oceans, characters, pregnancy
 &
also, health, may, used, one, disease, 
medical, use, first, people, 1, many, time, 
water, care\\
\hline
\texttt{GEN}&
gender, destruction, beliefs, schizophrenia, 
area, writers, armor, creativity, propagation, 
cheminformatics, overpopulation, deity,
stimulation, mathematical, cosmology & 
also, one, theory, 1, used, time, two, may, 
first, example, many, called, form, would, 
known\\
\hline
\texttt{GOV}& 
penology, tenure, governance, alloys,
 biosecurity, authoritarianism, criticisms, 
 burundi, motto, imperium, mesopotamia, 
 juche, 420, krytocracy, criticism &
also, government, political, state, united, 
party, one, minister, national,  states, 
first, would, used, new, university\\
  \hline
\texttt{LAW}&
alloys, biosecurity, authoritarianism, 
mesopotamia, electronic, economical, 
pupil, pathophysiology, imperium, phonology,
 collusion, cantons, auctoritas, sigint, juche & 
law, also, united, legal, may, act, states, 
court, rights, one, case, state, would, v, 
government\\ 
 \hline
 \texttt{LIF}&
 teacher, freight, career, agaricomycetes, 
 casein, manga, diplogasteria, benefit, 
 pteridophyta, basidiomycota, ascomycota, 
 letters, eukaryota, carcinogens, lifespan & 
also, used, may, often, one, made, water, 
food, many, use, usually, called,  known, 
oil, time\\
 \hline
 \texttt{MIL}& battles, eads, insignia, commanders,
 artillery, width, episodes, neurasthenia, 
 reconnaissance, elevation, freedom, length, 
 patrol, manufacturer, demise & 
military, war, army, also, air,  united, 
force, states,  one, used, forces, first, 
royal, british, world
 \\\hline
\end{tabular}
\caption{\label{tab:top} Top 15 domain words in the Wiki-en
  development set according to \wsdprior\ and L-LDA.}
\egroup
\end{table*}

Aside from automatic evaluation, we also assessed model performance
against human elicited domain labels for sentences and words.  The
purpose of this experiment was threefold: (a)~to validate the results
obtained from automatic evaluation; (b)~to evaluate finer-grained
model performance at the word level; and (c)~to examine whether our
model generalizes to non-Wikipedia articles. For this, we created a
third test set from the New York
Times\footnote{https://catalog.ldc.upenn.edu/LDC2008T19}, in addition
to our Wikipedia-based English and Chinese datasets. For all three
corpora, we randomly sampled two documents for each domain, and then
from each document, we sampled one long paragraph or a few consecutive
short paragraphs containing~\mbox{8--12} sentences.  Amazon Mechanical
Turkers were asked to read these sentences and assign a domain based
on the seven labels used in this paper (multiple labels were
allowed). Participants were provided with domain definitions. We
obtained five annotations per sentence and adopted the majority label
as the sentence's domain label. We obtained two annotated datasets for
English (Wiki-en and NYT-en) and one for Chinese (\mbox{Wiki-zh}),
consisting of 122/14, 111/11, and 117/12 sentences/documents each.

Word-level domain evaluation is more challenging; taken
out-of-context, individual words might be uninformative or carry
meanings compatible with multiple domains. Expecting crowdworkers to
annotate domain labels word-by-word with high confidence, might be
therefore problematic. In order to reduce annotation complexity, we
opted for a retrieval-style task for word evaluation. Specifically,
AMT workers were given a sentence and its domain label (obtained from
the sentence-level elicitation study described above), and asked to
highlight which words they considered consistent with the domain of
the sentence. We used the same corpora/sentences as in our first AMT
study.  Analogously, words in each sentence were annotated by five
participants and their labels were determined by majority agreement.

Fully hierarchical variants of our model (i.e.,~\wsd, \wsdprior) and L-LDA are
able to produce word-level predictions; we thus retrieved the words
within a sentence whose domain score was above the threshold of~0 and
compared them against the labels provided by
crowdworkers. \textsc{MilNet} and \sd\ can only make sentence-level
predictions. In this case, we assume that the sentence domain applies
to \emph{all} words therein. \textsc{HierNet} can
only produce document-level predictions based on which we generate
sentence labels and further assume that these apply to sentence words
too. 
Again, we report $\f$ which we compute as $\frac{2p^\ast
  r^\ast}{p^\ast+r^\ast}$ where precision $p^\ast$ and recall $r^\ast$
are both averaged over all words.

\begin{figure*}[t]
    \centering
   \includegraphics[width=16cm]{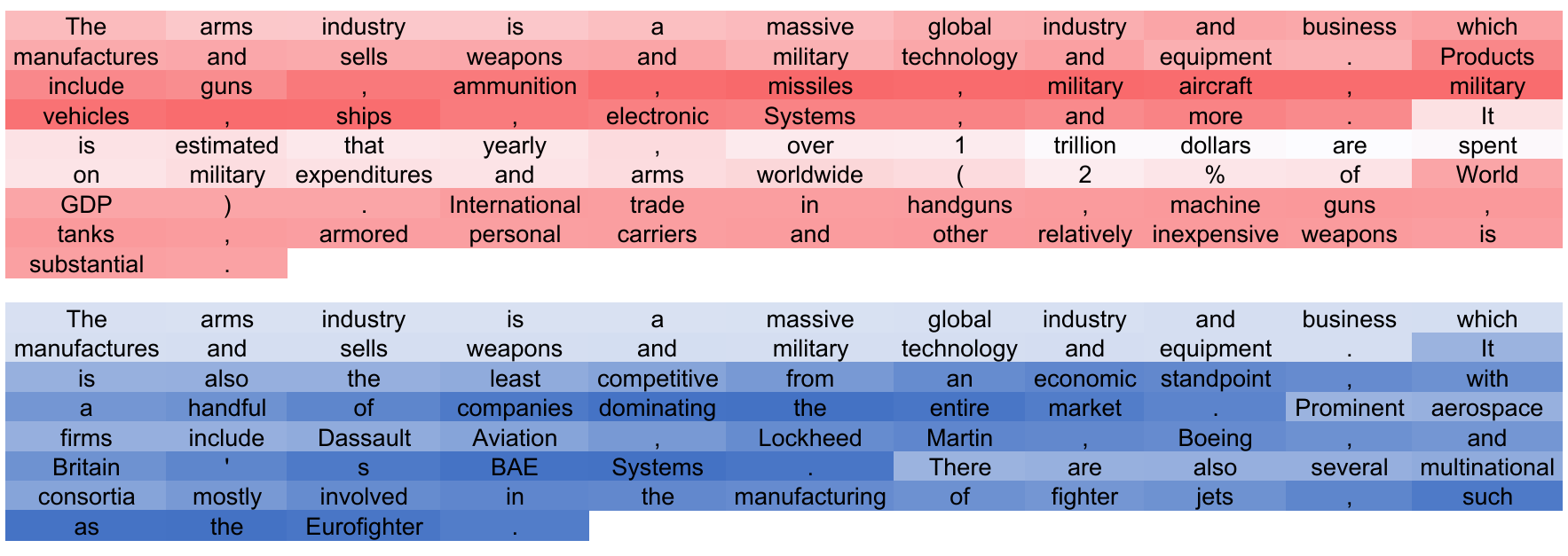}
   \caption{\label{fig:summary_en} Summaries for the Wikipedia article
     ``\textit{Arms Industry}''. The red heat map is for \texttt{MIL}
     and the blue one for \texttt{BUS}. Words with higher domain
     scores are highlighted with deeper color.}
 \end{figure*}

We show model performance against AMT domain labels in
Table~\ref{tab:human_eval}. Consistent with the automatic evaluation
results, \textsc{DetNet} variants are the best performing models on
the sentence-level task. On the Wikipedia datasets, \wsd~or
\wsdprior~outperform all baselines and \sd~by a large margin, showing
that word-level signals can indeed help detect sentence domains.
Although statistical models are typically less accurate when they are
applied to data that has a different distribution from the training
data, \wsdprior~works surprisingly well on NYT, substantially
outperforming all other systems. We also notice that prior information
is useful in making domain predictions for NYT sentences: since our
models are trained on Wikipedia, prior domain definitions largely
alleviate the genre shift to non-Wikipedia sentences.
Table~\ref{tab:performance_over_doms} provides a breakdown of the
performance of \wsdprior~across domains. Overall, the model performs
worst on \texttt{LIF} and \texttt{GEN} domains (which are very broad)
and best on \texttt{BUS} and \texttt{MIL} (which are very narrow).

With regard to word-level evaluation, \wsd\ and \wsdprior\ are the
best systems and are significantly better against all comparison
models by a wide margin, except L-LDA.  The latter is a strong domain
detection system at the word-level since it is able to \emph{directly}
associate words with domain labels (see
Equation~(\ref{eq:llda_count_mat})) without resorting to document- or
sentence-level predictions. However, our two-level hierarchical model
is superior considering all-around performance across sentences and
documents.  The results here accord with our intuition from previous
experiments: hierarchical models outperform simpler variants
(including \textsc{MilNet}) since they are able to capture and exploit
fine-grained domain signals relatively accurately. Interestingly,
prior information does not seem to have an effect on the Wikipedia
datasets, but is useful when transferring to NYT.  We also observe
that models trained on the Chinese datasets perform consistently
better than English. Analysis of the annotations provided by
crowdworkers revealed that the ratio of domain words in Chinese is
higher compared to English (27.47$\%$ vs. 13.86$\%$ in Wikipedia and
16.42$\%$ in NYT), possibly rendering word retrieval in Chinese an
easier task.



Table~\ref{tab:top} shows the 15 most representative domain words
identified by our model (\wsdprior) on Wiki-en for our seven
domains. We obtained this list by weighting word domain scores $\bm{P}$ with
their attention scores:
\begin{equation}
    \bm{P}^{\ast} = \bm{P} \odot [\bm{\alpha}; \ldots; \bm{\alpha}]^\intercal
\end{equation} 
and ranking all words in the development set according to
$\bm{P}^{\ast}$, separately for each domain. Since words appearing in
different contexts are usually associated with multiple domains, we
determine a word's ranking for a given domain based on the highest
score.  As shown in Table \ref{tab:top}, \textsl{biosecurity} and
\textsl{authoritarianism} are prevalent in both \texttt{GOV} and
\texttt{LAW} domains. Interestingly, with contextualized word
representations, fairly general English words are recognized as domain
heavy. For example, \textit{technique} is a strong domain word in
\texttt{HEA} and \textit{420} in \texttt{GOV} (the latter is slang for
the consumption of cannabis and highly associated with government
regulations).

For comparison, we also show the top domain words identified by L-LDA
via matrix~$\tilde{\bm{M}}$ (see
Equation~(\ref{eq:llda_count_mat})). To produce meaningful output, we
have removed stop words and punctuation tokens, which are given very
high domain scores by \mbox{L-LDA} (this is not entirely surprising
since~$\tilde{\bm{M}}$ is based on simple co-occurrence). Notice that
no such post-processing is necessary for our model.  As shown in
Table~\ref{tab:top}, the top domain words identified by L-LDA (on the
right) are more general and less informative, compared to those from
\wsdprior~(on the left).


\section{Domain-Specific Summarization}
\label{sec:summ}

In this section we illustrate how fine-grained domain scores can be
used to produce domain summaries, following an extractive,
unsupervised approach. We assume the user specifies the domains they
are interested in a priori (e.g., \texttt{LAW}, \texttt{HEA}) and the
system returns summaries targeting the semantics of these domains.

Specifically, we introduce \detrank, an extension of the well-known
\textrank~algorithm \cite{mihalcea2004textrank}, which incorporates
domain signals acquired by \wsdprior.  For each document,
\textrank~builds a directed graph $G=(V, E)$ with nodes~$V$
corresponding to sentences, and undirected edges~$E$ whose weights are
computed based on sentence similarity.  Specifically, edge weights are
represented with matrix $\bm{E}$ where each element $\bm{E}_{i, j}$
corresponds to the transition probability from vertex $i$ to vertex
$j$.  Following \newcite{barrios2016variations}, $\bm{E}_{i, j}$ is
computed with the Okapi BM25 algorithm \cite{robertson1995okapi}, a
probabilistic version of TF-IDF, and small weights ($<0.001$) are set
to zeros.  Unreachable nodes are further pruned to acquire the final
vertex set~$V$.

To enhance \textrank~with domain information, we first multiply
sentence-level domain scores $\bm{Q} $ with their corresponding attention
scores:
\begin{equation}
\label{eq:summ}
    \bm{Q}^{\ast} = \bm{Q} \odot [\bm{\beta}; \ldots; \bm{\beta}]^\intercal.
\end{equation} 
and for a given domain $c$, we can extract a (domain) sentence score
vector $\bm{q}^{\ast} = \bm{Q}^{\ast}_{c, *} \in \R^{1\times
  s}$. Then, from $\bm{q}^{\ast}$, we produce vector $\tilde{\bm{q}}
\in \R^{1\times |V|}$ representing a distribution of domain signals
over sentences:
\begin{gather}
	\hat{\bm{q}} = [\bm{q}^\ast_i]_{i \in V}\\
	\tilde{\bm{q}} = \softmax\left(\frac{\hat{\bm{q}} - \hat{\bm{q}}_{min}}{\hat{\bm{q}}_{max} - \hat{\bm{q}}_{min}}\right)
\end{gather}

In order to render domain signals in different sentences more
discernible,
we scale all elements in $\hat{\bm{q}}$ to $[0, 1]$ before obtaining a
legitimate distribution with the $\softmax$ function.  Finally, we
integrate the domain component into the original transition matrix as:
\begin{equation}
	\tilde{\bm{E}} = \phi * \hat{\bm{q}} + (1-\phi) * \bm{E} 
\end{equation} 
where $\phi \in (0, 1)$ controls the extent to which
domain-specific information influences sentence selection for the
summarization task; higher $\phi$ will lead to summaries which are
more domain-relevant.  Here, we empirically set $\phi=0.3$.  The main
difference between \detrank~and \textrank~is that \textrank~treats $1-
\phi$ as a damping factor and a uniform probability distribution is
applied to $\hat{\bm{q}}$.

In order to decide which sentence to include in the summary, a node’s
centrality is measured using a graph-based ranking algorithm
\cite{mihalcea2004textrank}. Specifically, we run a Markov chain
with~$\tilde{\bm{E}}$ on $G$ until it converges to the stationary
distribution $\bm{e}^\ast$ where each element denotes the salience of
a sentence.  In the proposed \detrank\ algorithm, $\bm{e}^\ast$
jointly expresses the importance of a sentence in the document
\emph{and} its relevance to the given domain (controlled by $\phi$).
We rank sentences according to $\bm{e}^\ast$ and select the top~$K$
ones, subject to a budget (e.g.,~100 words).

We ran a judgment elicitation study on summaries produced by
\textrank~and \detrank.  Participants were provided with domain
definitions and asked to decide which summary was best according to
the criteria of: \textit{Informativeness} (does the summary contain
more information about a specific domain, e.g., ``Government and
Politics''?), \textit{Succinctness} (does the summary avoid
unnecessary detail and redundant information?), and \textit{Coherence}
(does the summary make logical sense?). Amazon Mechanical Turk (AMT)
workers were allowed to answer ``Both'' or ``Neither'' in cases where
they could not discriminate between summaries.  We sampled 50 summary
pairs from the English Wikipedia development set. We collected three
responses per summary pair and determined which system participants
preferred based on majority agreement.

\begin{table}[t]
\small
\centering
\tabcolsep=0.05cm
\begin{tabular}{|l|cccc|} \hline
\multicolumn{1}{|c|}{\textbf{Method}} & \textbf{Inf} & \textbf{Succ} & \textbf{Coh} & \textbf{All} \\\hline\hline
\textrank & \hspace*{1ex}$45.45^\dagger$     & \hspace{.1ex}$\bm{51.11}$ & \hspace*{.8ex}$42.50\dagger$ & \hspace*{.8ex}$46.35\dagger$\\
\detrank & $\bm{54.55}$ & $48.89$      & $\bm{57.50}$ & $\bm{53.65}$\\
\hline
\end{tabular}
\caption{\label{tab:summ_results} Human evaluation results for 
  summaries produced by \textrank~and \detrank; proportion of
  times AMT workers found models Informative (Inf), Succinct (Succ),
  and Coherent (Coh); All is the average across ratings;  symbol~$\dagger$
  denotes that differences between models are statistically significant 
  (\mbox{$p < 0.05$}) using a pairwise t-test.} 
\end{table}

Table~\ref{tab:summ_results} shows the proportion of times AMT workers
preferred each system according to the criteria of Informativeness,
Succinctness, Coherence, and overall. As can be seen, participants
find \detrank~summaries more informative and coherent.  While it is
perhaps not surprising for \detrank~to produce summaries which are
domain informative since it explicitly takes domain signals into
account, it is interesting to note that focusing on a specific domain
also helps discard irrelevant information and produce more coherent
summaries.  This, on the other hand, possibly renders \detrank's
summaries more verbose (see the Succinctness ratings in
Table~\ref{tab:summ_results}).

Figure~\ref{fig:summary_en} shows example summaries for the Wikipedia
article \textit{Arms Industry} for domains \texttt{MIL} and
\texttt{BUS}.\footnote{https://en.wikipedia.org/wiki/Arms\_industry}
Both summaries begin with a sentence which introduces the arms
industry to the reader. When \texttt{MIL} is the domain of interest,
the summary focuses on military products such as guns and missiles.
When the domain changes to \texttt{BUS}, the summary puts more
emphasis on trade, e.g.,~market competition and companies doing
military business, such as Boeing and Eurofighter.


\section{Conclusions}
\label{sec:conclusions}

In this work, we proposed an encoder-detector framework for domain
detection. Leveraging only weak domain supervision, our model achieves
results superior to competitive baselines across different languages,
segment granularities, and text genres. Aside from identifying domain
specific training data, we also show that our model holds promise for
other natural language tasks, such as text summarization.  Beyond
domain detection, we hope that some of the work described here might
be of relevance to other multilabel classification problems such as
sentiment analysis \cite{angelidis2018multiple}, relation extraction
\cite{surdeanu-EtAl:2012:EMNLP-CoNLL}, and named entity recognition
\cite{tang-EtAl:2017:EMNLP2017}. More generally, our experiments show
that the proposed framework can be applied to textual data using
minimal supervision, significantly alleviating the annotation
bottleneck for text classification problems. 

A key feature in achieving performance superior to competitive
baselines is the hierarchical nature of our model, where
representations are encoded step-by-step, first for words, then for
sentences, and finally for documents. The framework flexibly
integrates prior information which can be used to enhance the
otherwise weak supervision signal or to render the model more robust
across genres. In the future, we would like to investigate
semi-supervised instantiations of MIL, where aside from bag labels,
small amounts of instance labels are also available
\cite{kotzias2015group}.  It would also be interesting to examine how
the label space influences model performance, especially since in our
scenario the labels are extrapolated from Wikipedia and might be naturally
noisy and/or ambiguous.

\section*{Acknowledgments}
The authors would like to thank the anonymous reviewers and the action
editor, Yusuke Miyao, for their valuable feedback. We acknowledge the
financial support of the European Research Council (Lapata; award
number 681760). This research is based upon work supported in part by
the Office of the Director of National Intelligence (ODNI),
Intelligence Advanced Research Projects Activity (IARPA), via contract
FA8650-17-C-9118.  The views and conclusions contained herein are
those of the authors and should not be interpreted as necessarily
representing the official policies or endorsements, either expressed
or implied, of the ODNI, IARPA, or the U.S. Government. The
U.S. Government is authorized to reproduce and distribute reprints for
Governmental purposes notwithstanding any copyright annotation
therein.


\bibliographystyle{acl_natbib}

\end{document}